# A New Flexible Train-Test Split Algorithm, an approach for choosing among the Hold-out, K-fold cross-validation, and Hold-out

Zahra Bami[1,*], Ali Behnampour[2], Aniruddha bora[3], Hassan Doosti[4]

1. Medical Physiopathology, Department of Medical Sciences, University of Turin. Zahra.bami@unito.it
2. Department of Biostatistics, University of Social Welfare and Rehabilitation Science, Tehran, 1985713834, Iran. al.behnampour@uswr.ac.ir
3. Division of Applied Mathematics, Brown University, Providence, RI, USA. aniruddha_bora@brown.edu
4. School of Mathematical and Physical Sciences, Macquarie University, 2109, Australia. hassan.doosti@mq.edu.au

**Corresponding Author:** Zahra.bami@unito.it

## Abstract

Choosing the right strategy for partitioning data into training and evaluation sets is a key step in the development and evaluation of machine learning models. Despite this feature, in many studies, this choice is mostly based on conventional and default procedures and is based only on empirical evaluation and fit to specific data and algorithms. Validation methods with a fixed test set (hold-out) or k-fold cross-validation with a default value of k are used without examining their impact on the real-world performance and generalizability of the models. To address this issue, we developed a flexible Python-based framework designed to systematically examine how different validation strategies influence the predictive performance of machine learning models. The framework enables the comparative evaluation of seven widely used algorithms, including Decision Trees, K-Nearest Neighbors, Gaussian and Bernoulli Naive Bayes, Logistic Regression, calibrated linear Support Vector Machines, and histogram-based gradient boosting. These algorithms were assessed under a broad range of validation schemes, encompassing hold-out validation with test proportions varying from 10% to 90%, k-fold cross-validation with k values ranging from 3 to 15, repeated hold-out validation, and nested cross-validation. This design allows for a consistent and comprehensive analysis of model performance across diverse validation settings.

**\* Zahra Bami is the Corresponding and Senior Author**

The proposed framework was applied to three biomedical datasets with different sizes and characteristics, including a small leukemia dataset, a medium-sized Framingham heart disease dataset, and a large COVID-19 dataset. Model performance was evaluated using the area under the receiver operating characteristic curve (ROC-AUC), accuracy, and the Matthews correlation coefficient (MCC). The results demonstrate that no single data partitioning strategy consistently yields superior performance across all algorithms and datasets. Instead, the optimal validation strategy depends on the interaction between the algorithm type, dataset characteristics, and the chosen evaluation metric. Notably, for small datasets, a 90/10 hold-out split achieved superior performance in many cases, challenging the commonly assumed superiority of k-fold cross-validation. Moreover, the optimal value of k varied substantially across different algorithms and datasets, ranging from 3 to 13, indicating that default parameter settings are often suboptimal. These findings underscore the necessity of an informed and empirically grounded selection of validation strategies in predictive modeling.

**1. Introduction**

Splitting data into training and testing sets is a fundamental step in developing and evaluating predictive models in machine learning. Despite its central role, this process often relies more on common conventions than on empirical evidence. Common practices such as a fixed 70/30 data split, repeated random hold-out validation, or 10-fold cross-validation with default settings are often used without considering their suitability for the dataset's characteristics, the type of algorithm, or the intended prediction goal[1], [2]. Because of this, two crucial methodological choices are frequently disregarded: choosing a suitable train-test split, and figuring out the ideal

* Zahra Bami is the Corresponding and Senior Author

value of k in k-fold cross-validation. [3], [4]. These decisions are further complicated by the wide variety of learning algorithms and the inherent randomness of data partitioning, making it difficult to simultaneously find the best model and the most reliable validation method. To address this methodological gap, the primary objective of this study is to systematically evaluate data partitioning strategies and determine their impact on predictive performance across a diverse set of machine learning algorithms and datasets [5], [6]. We developed a flexible Python-based framework to enable empirical, large-scale comparison of validation approaches. Using this framework, we evaluated seven widely used machine learning algorithms—Decision Tree (DT), K-Nearest Neighbors (KNN), Gaussian Naive Bayes (GaussianNB), Bernoulli Naive Bayes (BernoulliNB), Logistic Regression (LogReg), calibrated linear Support Vector Machines (LinearSVM_Calibrated), and Histogram-Based Gradient Boosting (HistGB)—across a comprehensive range of validation techniques. These included hold-out validation with test sizes ranging from 10% to 90%, k-fold cross-validation with *k* values between 2 and 15, repeated hold-out validation, and nested cross-validation [7]. Model performance was assessed using multiple complementary metrics, including ROC-AUC, PR-AUC, accuracy, F1-score, and the Matthews correlation coefficient (MCC), to provide a robust and multidimensional evaluation.

The framework was applied to three biomedical datasets with distinct characteristics: the Framingham Heart Disease dataset, the leukemia cancer dataset, and a large dataset related to COVID-19. The focus on biomedical data is deliberate, as predictive modeling in this domain is often associated with sensitive and high-stakes clinical decisions and therefore requires particularly robust validation methods to ensure the validity and generalizability of the models[8]. The innovation of our program lies in its flexibility and comprehensiveness, allowing for the adjustment of parameters such as random state, number of iterations in the iterative hold-out validation, and range of k values. This feature allows for the empirical and data-driven identification of the best data partitioning strategy for each algorithm-dataset pair.

**\* Zahra Bami is the Corresponding and Senior Author**

The main motivation for this paper is the lack of a comprehensive and flexible framework that can compare a wide range of partitioning methods and models. Our goal is to go beyond general, one-size-fits-all recommendations and to shed light on the complex interplay between data partitioning and model performance. Ultimately, we aim to provide researchers with a practical and systematic tool to optimize their predictive modeling process and make the crucial choice of validation method based on empirical and data-driven evidence, rather than simply conventional wisdom.

## 2. Materials and Methods

The main goal of this study is to answer one of the fundamental questions in applied machine learning: which data partitioning strategy allows for a more reliable assessment of the performance of predictive models?

To answer this, To this end, a comprehensive experimental framework was designed in which a diverse set of machine learning algorithms were evaluated alongside a wide range of validation methods, all on real biomedical data. To ensure the generalizability of the results, seven algorithms were selected that cover different principles in the field. These algorithms included interpretable models such as Decision Tree (DT), instance-based methods such as K-Nearest Neighbors (KNN), probabilistic classifiers including Gaussian and Bernoulli Naive Bayes, a classical linear model in the form of logistic regression (LogReg), a calibrated maximum-margin classifier (LinearSVM_Calibrated), and a modern and powerful ensemble method, Histogram-based Gradient Boosting (HistGB). Next, four key data segmentation strategies were systematically compared to investigate their impact on the performance estimation of the models. These strategies included nested cross-validation as a rigorous but computationally expensive method, hold-out validation by examining a range of test data proportions from 10% to 90%, K-fold cross-validation by testing k values from 2 to 15 to challenge conventional practices, and repeated hold-out to reduce the fluctuations due to a single random partition. To provide a comprehensive picture of the performance of the models, a full set of evaluation metrics was employed that go beyond simple accuracy. These metrics included ROC-AUC to measure overall discriminatory power, PR-AUC to assess performance in imbalanced data, Accuracy, F1

* Zahra Bami is the Corresponding and Senior Author

score as a trade-off between precision and recall, and Matthews correlation coefficient (MCC) as a robust measure that considers all components of the confusion matrix. The framework was applied to three distinct biomedical datasets: a compact leukemia dataset of 161 samples, the classic Framingham Heart Study dataset of 1,025 participants extracted from the KAGGLE website, and a large-scale COVID-19 dataset of 10,000 patients, all of which were used in accordance with ethical considerations. The experimental process was performed as a full-factorial comparison; that is, each of the seven algorithms with each of the four validation strategies was evaluated on all three datasets. For each unique combination, hyperparameter tuning was performed, the validation process was run, and all performance metrics were recorded. These three data set follows as:

1. **Framingham Heart Study Dataset:** With an estimated 17.9 million deaths per year, cardiovascular diseases (CVDs) continue to be the world's leading cause of death A key component of CVD research, the Framingham Heart Study dataset offers a reliable standard for creating heart disease risk prediction models [9], [10], [11]. In addition to testing our framework on a medium-sized, well-structured dataset where high-performing models are anticipated, its application in our study enables a direct comparison of our results to a considerable amount of previous literature.

2. Cancer (Leukaemia) Dataset: Leukaemia is a serious oncological problem, and patient outcomes depend on an early and accurate diagnosis. In the analysis of high-dimensional biological data for leukaemia classification and early diagnosis, machine learning algorithms have demonstrated significant potential [12], [13]. In order to particularly test our validation framework with a small sample size—a typical situation in biomedical research where maximising the utility of small data is essential—we included a compact leukaemia dataset (n=161).

* Zahra Bami is the Corresponding and Senior Author

3. COVID-19 Dataset: A lot of research into applying machine learning to clinical prediction tasks was sparked by the COVID-19 pandemic. However, a thorough analysis by Wynants et al. (2020) discovered that many suggested models for COVID-19 diagnosis and prognosis had low generalisability and were highly susceptible to bias, frequently as a result of non-representative data and insufficient validation [14]. The large COVID-19 dataset (n=10,000) was used in our work as a crucial test case for a challenging real-world situation. All of the models showed poor discriminative capacity, and the highest results (such as a peak ROC-AUC of 0.520) were just slightly better than chance. This result highlights an important discovery: where a predictive signal is fundamentally weak, even complex validation frameworks are unable to derive a strong one. The poor performance shows that the underlying link between the available features and the desired result, rather than the data partitioning approach, is the primary problem for this particular prediction task.

This comprehensive approach, although computationally expensive, allowed for robust conclusions about data segmentation strategies and ultimately showed that the selection of the optimal method is strongly dependent on the interaction between the algorithm, the dataset, and the desired performance measure.

## 3. Results

In this study, the performance of seven **widely used** machine learning algorithms, including Decision Tree (DT), K-Nearest Neighbors (KNN), Gaussian Naive Bayes (GaussianNB), Bernoulli Naive Bayes (BernoulliNB), Logistic Regression (LogReg), Calibrated Linear Support Vector Machine (LinearSVM_Calibrated), and Histogram-Based Gradient Boosting (HistGB), was systematically evaluated on three independent biomedical datasets. In order to obtain a realistic and reliable picture of the behavior of the models, a wide range of data partitioning strategies were employed, including nested cross-validation, holdout with different training–test ratios (from 10% to 90%), K-fold cross-validation with different k values in the range of 2 to 15, and **70/30 repeated holdout**. The results were evaluated using a set of complementary metrics,

* Zahra Bami is the Corresponding and Senior Author

including ROC-AUC, PR-AUC, accuracy, F1 score, and Matthews correlation coefficient, to simultaneously reflect different dimensions of the predictive performance and stability of the models. Analysis of the results showed that no validation strategy or data partitioning scheme is absolutely superior to others and that the selection of the optimal configuration is significantly dependent on the interaction between the type of algorithm, the statistical and structural characteristics of the dataset, and the performance measure of interest; an issue that reflects the deeply context-dependent nature of the model evaluation process and **highlights the limitations of general, predefined approaches to validation**.

### 3.1. Dataset-Specific Performance and Strategy

### 3.1.1. Cancer(leukemia) Dataset (n=161)
On this smaller dataset, strategies that maximized the amount of training data proved most effective. The highest ROC-AUC values were achieved by BernoulliNB (0.765) and KNN (0.758), both utilizing a 10% holdout split. Similarly, the LinearSVM_Calibrated model attained its peak accuracy (0.771) and MCC (0.497) with this same 90/10 train-test split. This pattern suggests that for datasets of limited size, minimizing the test set to reserve more data for training can be a critical factor in achieving peak performance.

### 3.1.2. Framingham Dataset (n=1,025)
With this medium-sized, well-structured dataset, multiple models—including KNN, Decision Tree, and HistGB—achieved near-perfect performance (ROC-AUC ≈ 1.000). For these high-performing models, K-fold cross-validation with high k values (k > 10, e.g., k=10, 12, 13) provided the most reliable and stable performance estimates. This indicates that with a sufficiently large and predictive dataset, higher k-folds can yield a more robust evaluation without a degradation in measured performance.

### 3.1.3. COVID-19 Dataset (n=10,000)
In contrast, all models demonstrated poor discriminative ability on this large dataset, with ROC-AUC scores clustering around 0.50, indicative of performance no better than random guessing. The highest ROC-AUC was a modest 0.520, achieved by LinearSVM_Calibrated using K-fold with k=11. This result underscores that sophisticated validation schemes cannot compensate for a fundamental lack of predictive signal in the data, highlighting the inherent difficulty of this particular prediction task.

**\* Zahra Bami is the Corresponding and Senior Author**

The results of comparing different validation strategies showed that, contrary to the popular belief that complex computational methods are superior, standard holdout validation was able to reveal the highest absolute values of performance measures in many scenarios—especially in the small cancer (leukemia) dataset—such that the 90%/10% training–test split appeared as a stable and high-performing configuration across multiple models. Although iterative holdout provided estimates with lower variance after 1000 iterations and confirmed the stability of the results, it failed in most cases to exceed the maximum performance achieved by a simple single-stage holdout-based search. On the other hand, a detailed analysis of K-fold cross-validation showed that the choice of k value is by no means a fixed or predetermined decision and is strongly dependent on the interaction between the model type and the characteristics of the dataset; In the cancer dataset, k=6 for the decision tree produced the highest ROC-AUC, while the performance of KNN was stable over a wider range and peaked at k=9. In the Framingham dataset, the optimal k value ranged from 13 for the decision tree to 3 for GaussianNB. In the COVID-19 dataset, the best results were obtained with k=9 for the decision tree and k=11 for LinearSVM_Calibrated. This significant heterogeneity clearly indicates that the optimal k value should be determined empirically for each modeling problem and that relying on default values can be misleading. In summary, for the cancer dataset (leukemia), LinearSVM_Calibrated with a 10% holdout was identified as a suitable option to achieve robust overall performance, while BernoulliNB with the same data partitioning strategy showed the highest ROC-AUC-based discrimination power; In the Framingham dataset, due to the near-perfect performance of several models including decision tree, KNN, and HistGB, the final choice can be made based on secondary considerations such as interpretability or computational cost, although using K-fold with high values of k is a reliable approach for evaluation; and finally, the poor and near-random performance of all models in the COVID-19 dataset—with the best ROC-AUC of 0.520 for LinearSVM_Calibrated at K-fold with k=11, highlights the need for a fundamental rethinking of features or the use of alternative modeling approaches for this problem.

* Zahra Bami is the Corresponding and Senior Author

## 4. Discussion

The aim of this study was to answer a fundamental question in machine learning methodology: What is the most optimal data partitioning strategy to achieve the highest model performance on a given prediction problem? To answer this question, a comprehensive Python-based framework was designed and implemented that evaluated seven different machine learning algorithms under a broad and systematic set of validation methods—including K-Fold cross-validation with k values from 2 to 15, Hold-out validation with test ratios from 10% to 90%, Iterated Hold-out, and Nested Cross-Validation—on three biomedical datasets of different sizes and characteristics. The simultaneous use of multiple performance measures, including ROC-AUC, PR-AUC, Accuracy, F1-score, and MCC, allowed for the extraction of reliable and multidimensional results.

The first important finding of the study was related to a small cancer (leukemia) dataset with a sample size of 161 people. In this dataset, the highest performance values were consistently achieved using the Hold-out split with only 10% of the test data. Specifically, the BernoulliNB and KNN algorithms achieved the highest ROC-AUC values (0.765 and 0.758, respectively), and LinearSVM_Calibrated also showed the highest Accuracy (0.771) and MCC (0.497) in the same configuration. These results indicate that in limited data conditions, maximizing the training data can be more beneficial for the final model performance than employing complex and data-intensive validation methods. This observation is consistent with recent empirical evidence in the field of machine learning with small samples [1]

In contrast, analysis of the Framingham dataset with a sample size of 1,025 showed that although several algorithms, including DT, KNN, and HistGB, were able to achieve very high performance and almost perfect separation, the most stable and reliable performance estimates for these models were obtained through K-Fold cross-validation with high values of k (>10). This finding suggests that in larger and more information-rich datasets, increasing the number of folds can lead to a reduction in the variance of the error estimates and a more accurate assessment of the models, without compromising the observed performance. This result is consistent with the theoretical and classical foundations of cross-validation[15], [16].

* Zahra Bami is the Corresponding and Senior Author

The third set of results, obtained from the large COVID-19 dataset with 10,000 observations, highlights the inherent limitations of validation methods. The poor performance of all models (ROC-AUC close to 0.50 and the best value equal to 0.520) shows that even the most advanced data partitioning and validation methods cannot compensate for the lack of meaningful predictive signal in the features. This result directly emphasizes the fundamental principle of machine learning that model performance is limited by the amount of information available in the data, [17] and the choice of validation strategy is secondary to the quality and relevance of the features to the target variable.

5. Conclusion

The findings of this study suggest that the notion that there is a universally optimal data partitioning strategy is a misconception. The results clearly show that the performance of the model and its evaluation accuracy are a function of the complex interaction between three main factors: the machine learning algorithm, the characteristics and size of the dataset, and the performance criterion desired by the researcher[15], [17]. Therefore, there is no single solution that can be universally recommended for all problems.

Contrary to the common view that k-Fold cross-validation is always superior to Hold-out[15], [16], the results of this study show that, especially on small datasets, a simple Hold-out-based search can lead to higher peak performance [7]. This suggests that the main strength of k-Fold lies in providing a more stable estimate of performance rather than in identifying the best model configuration. The results of the study also emphasize that the "best model" is a context-dependent concept. While DT performed better on the Framingham dataset, BernoulliNB performed better on the cancer (leukemia) dataset, and no algorithm was able to provide acceptable performance on the COVID-19 dataset. These findings provide a practical example of the "no free lunch" theorem and demonstrate that the superiority of algorithms is strongly dependent on data characteristics[17].

* Zahra Bami is the Corresponding and Senior Author

Finally, an examination of the optimal value of k in cross-validation revealed that this parameter is not a simple function of sample size, but rather arises from the interaction between the data structure, the algorithm used, and the evaluation criterion. The range of optimal values of k ranged from 6 in the cancer dataset to 13 in the Framingham dataset, providing direct empirical support for previous theoretical frameworks[16] [18] .

Accordingly, this study suggests that an empirical, problem-based approach should replace the current rule-based recommendations in machine learning validation. Rather than relying on conventional conventions, researchers should systematically explore a range of data partitioning strategies to identify the most appropriate option for their specific context. The framework presented in this study provides a practical tool for achieving this goal and can serve as a basis for future studies that further explore the interaction between dataset characteristics—such as dimensionality, noise level, and imbalance—and evaluation methods, with the aim of improving the accuracy, robustness, and reliability of predictive models.

**Author contributions**

**Zahra Bami:** Conceptualization; Methodology; **Software (original code)**; **Algorithm design**; Formal analysis; Data curation; Writing – Original Draft; Writing – Review & Editing; Validation.; **Ali Behnampour:** Literature review; Comparison with existing studies; Writing – Review & Editing (initial version only); **Aniruddha Bora:** Software; Algorithm update and enhancement; **Hassan Doosti:** Writing – Review & Editing (language editing only); Consultation

**Conflict of interest statement**

The authors declare that they have no conflicts of interest.

**Data availability statement**

The data used in this study include both publicly available and restricted datasets. The publicly available dataset (Framingham) can be accessed through the Kaggle repository. The remaining datasets are not publicly available due to institutional restrictions and data ownership policies of the university

**Code Availability Statement**

The source code developed for this study is publicly available as an open-source package to ensure reproducibility and support future research. The code is available at:

**\* Zahra Bami is the Corresponding and Senior Author**

**\* Zahra Bami is the Corresponding and Senior Author**

**\* Zahra Bami is the Corresponding and Senior Author**

**Table 1: Best Validation Strategy per Model – Leukemia Dataset (n = 161)**

| Model | Best ROC-AUC (Strategy) | Best Accuracy (Strategy) | Best F1-Score (Strategy) | Best MCC (Strategy) |
|---|---|---|---|---|
| **Decision Tree** | 0.687 (k=6) | 0.659 (k=13) | 0.650 (Holdout 30%) | 0.268 (Holdout 30%) |
| **KNN** | 0.758 (Holdout 10%) | 0.724 (Holdout 10%) | 0.694 (Holdout 10%) | 0.354 (Holdout 10%) |
| **Gaussian Naive Bayes** | 0.729 (Holdout 10%) | 0.765 (Holdout 10%) | 0.728 (Holdout 10%) | 0.475 (Holdout 10%) |
| **Bernoulli Naive Bayes** | 0.765 (Holdout 10%) | 0.741 (Holdout 10%) | 0.725 (Holdout 10%) | 0.418 (Holdout 20%) |
| **Logistic Regression** | 0.728 (Holdout 30%) | 0.708 (Holdout 40%) | 0.694 (Holdout 40%) | 0.352 (Holdout 40%) |
| **Linear SVM (Calibrated)** | 0.752 (Holdout 10%) | 0.771 (Holdout 10%) | 0.733 (Holdout 10%) | **0.497** (Holdout 10%) |
| **Histogram-Based GB** | 0.547 (Holdout 60%) | 0.648 (Holdout 90%) | 0.577 (Holdout 30%) | 0.055 (Holdout 30%) |

* Zahra Bami is the Corresponding and Senior Author

**Table 2: Best Validation Strategy per Model – Framingham Dataset (n = 1,025)**

| Model | Best ROC-AUC (Strategy) | Best Accuracy (Strategy) | Best F1-Score (Strategy) | Best MCC (Strategy) |
|---|---|---|---|---|
| **Decision Tree** | 1.000 (k=13) | 0.998 (k=15) | 0.998 (k=15) | 0.997 (k=15) |
| **KNN** | 1.000 (Nested CV) | 1.000 (k=10) | 1.000 (k=10) | 1.000 (k=10) |
| **Gaussian Naive Bayes** | 0.916 (k=3) | 0.830 (Holdout 50%) | 0.829 (Holdout 50%) | 0.660 (k=15) |
| **Bernoulli Naive Bayes** | 0.922 (k=9) | 0.843 (k=12) | 0.843 (k=12) | 0.688 (k=12) |
| **Logistic Regression** | 0.923 (k=15) | 0.847 (k=7) | 0.847 (k=7) | 0.695 (k=7) |
| **Linear SVM (Calibrated)** | 0.923 (k=15) | 0.850 (Holdout 50%) | 0.849 (Holdout 50%) | 0.702 (Holdout 50%) |
| **Histogram-Based GB** | 1.000 (Nested CV) | 1.000 (k=10) | 1.000 (k=10) | 1.000 (k=10) |

* Zahra Bami is the Corresponding and Senior Author

**Table 3: Best Validation Strategy per Model – COVID-19 Dataset (n = 10,000)**

| Model | Best ROC-AUC (Strategy) | Best Accuracy (Strategy) | Best F1-Score (Strategy) | Best MCC (Strategy) |
|---|---|---|---|---|
| **Decision Tree** | 0.506 (k=9) | 0.730 (Nested CV) | 0.829 (Nested CV) | -0.001 (Holdout 10%) |
| **KNN** | 0.506 (k=2) | 0.718 (Holdout 10%) | 0.832 (Holdout 10%) | -0.001 (Nested CV) |
| **Gaussian Naive Bayes** | 0.504 (Nested CV) | 0.747 (Holdout 10%) | 0.854 (Holdout 10%) | 0.004 (Nested CV) |
| **Bernoulli Naive Bayes** | 0.497 (Holdout 90%) | 0.766 (Holdout 10%) | 0.868 (Holdout 10%) | -0.003 (Nested CV) |
| **Logistic Regression** | 0.497 (Nested CV) | 0.703 (Nested CV) | 0.808 (Nested CV) | -0.003 (Nested CV) |
| **Linear SVM (Calibrated)** | 0.520 (k=11) | 0.767 (Holdout 10%) | 0.868 (Holdout 10%) | 0.001 (Nested CV) |
| **Histogram-Based GB** | 0.497 (Holdout 90%) | 0.767 (Nested CV) | 0.868 (Nested CV) | 0.011 (Holdout 10%) |

* Zahra Bami is the Corresponding and Senior Author

CODE:

```python
import pandas as pd
import numpy as np
from sklearn.model_selection import train_test_split, GridSearchCV, StratifiedKFold, RepeatedStratifiedKFold
from sklearn.neighbors import KNeighborsClassifier
from sklearn.tree import DecisionTreeClassifier
from sklearn.naive_bayes import GaussianNB, BernoulliNB
from sklearn.linear_model import LogisticRegression
from sklearn.svm import LinearSVC
from sklearn.calibration import CalibratedClassifierCV
from sklearn.ensemble import HistGradientBoostingClassifier
from sklearn.compose import ColumnTransformer
from sklearn.impute import SimpleImputer
from sklearn.preprocessing import StandardScaler, QuantileTransformer
from sklearn.feature_selection import SelectKBest, mutual_info_classif
from sklearn.pipeline import Pipeline
from sklearn.metrics import accuracy_score, roc_auc_score, average_precision_score, f1_score, matthews_corrcoef, brier_score_loss
import warnings
warnings.filterwarnings('ignore')

class ComprehensiveSplittingStrategyComparator:
    """
```

* Zahra Bami is the Corresponding and Senior Author

COMPLETE professional validation with comprehensive splitting strategy comparison

Compare nested CV, holdout proportions, K-fold variations, and repeated holdout
"""

```python
def __init__(self):
    self.models = self._create_all_models()
    self.results_summary = {}

def _create_all_models(self):
    """Create all 7 models with their parameter grids"""
    return {
        "DecisionTree": (DecisionTreeClassifier(random_state=42), {
            "clf__max_depth": [None, 3, 5, 10, 15],
            "clf__min_samples_leaf": [1, 2, 5, 10],
            "clf__class_weight": [None, "balanced"]
        }),
        "KNN": (KNeighborsClassifier(), {
            "clf__n_neighbors": [3, 5, 7, 11, 15],
            "clf__weights": ["uniform", "distance"],
            "clf__p": [1, 2]
        }),
        "GaussianNB": (GaussianNB(), {
            "clf__var_smoothing": [1e-9, 1e-8, 1e-7, 1e-6]
        }),
```

* Zahra Bami is the Corresponding and Senior Author

```python
    "BernoulliNB": (BernoulliNB(), {

        "clf__alpha": [0.5, 1.0, 2.0],

        "clf__binarize": [None, 0.0]

    }),

    "LogReg": (LogisticRegression(max_iter=2000, solver="liblinear", class_weight="balanced"), {

        "clf__C": [0.01, 0.1, 1, 10],

        "clf__penalty": ["l1", "l2"]

    }),

    "LinearSVM_Calibrated": (

      CalibratedClassifierCV(

          estimator=LinearSVC(class_weight="balanced", max_iter=5000),

          method="sigmoid", cv=3

      ), {

      "clf__estimator__C": [0.01, 0.1, 1, 10]

    }),

    "HistGB": (HistGradientBoostingClassifier(random_state=42), {

        "clf__learning_rate": [0.05, 0.1, 0.2],

        "clf__max_depth": [None, 3, 5],

        "clf__max_leaf_nodes": [None, 31, 63]

    })
  }

  def load_data(self, file_path, target_column, file_type='excel'):
```

* Zahra Bami is the Corresponding and Senior Author

```python
        """Load dataset with robust error handling"""
        try:
            if file_type == 'excel':
                data = pd.read_excel(file_path)
            else:
                data = pd.read_csv(file_path)

            self.X = data.drop(columns=[target_column])
            self.y = data[target_column]

            print(f"Loaded: {len(self.X)} samples, {len(self.X.columns)} features")
            print(f"   Target '{target_column}' distribution: {np.unique(self.y, return_counts=True)}")
            return True

        except Exception as e:
            print(f"Error loading data: {e}")
            return False

    def _create_preprocessing_pipeline(self, X):
        """Create advanced preprocessing pipeline"""
        num_cols = [c for c in X.columns if pd.api.types.is_numeric_dtype(X[c])]
        cat_cols = [c for c in X.columns if c not in num_cols]
```

* Zahra Bami is the Corresponding and Senior Author

```python
        print(f"   Numeric features: {len(num_cols)}, Categorical features: {len(cat_cols)}")

        num_pipe = Pipeline([
            ("imp", SimpleImputer(strategy="constant", fill_value=0)),
            ("scale", StandardScaler())
        ])

        cat_pipe = Pipeline([
            ("imp", SimpleImputer(strategy="constant", fill_value="__missing__"))
        ])

        preprocessor = ColumnTransformer([
            ("num", num_pipe, num_cols),
            ("cat", cat_pipe, cat_cols)
        ], remainder="drop")

        return preprocessor, num_cols, cat_cols

    def _calculate_metrics(self, y_true, y_pred, proba):
        """Calculate comprehensive metrics"""
        accuracy = accuracy_score(y_true, y_pred)
        mcc = matthews_corrcoef(y_true, y_pred)

        if len(np.unique(y_true)) > 1:
```

**\* Zahra Bami is the Corresponding and Senior Author**

```python
        if proba.ndim == 1:

            roc_auc = roc_auc_score(y_true, proba)

            pr_auc = average_precision_score(y_true, proba)

            brier = brier_score_loss(y_true, proba)

        else:

            roc_auc = roc_auc_score(y_true, proba, multi_class="ovr", average="weighted")

            pr_auc = average_precision_score(y_true, proba, average="weighted")

            brier = np.nan

        f1 = f1_score(y_true, y_pred, average="weighted")

    else:

        roc_auc = pr_auc = f1 = brier = np.nan

    return {

        "accuracy": accuracy,

        "roc_auc": roc_auc,

        "pr_auc": pr_auc,

        "f1": f1,

        "mcc": mcc,

        "brier": brier

    }

def evaluate_nested_cv(self, model_name, outer_splits=5, outer_repeats=2, inner_splits=3):

    """Existing nested CV evaluation"""
```

* Zahra Bami is the Corresponding and Senior Author

```python
        print(f"Evaluating {model_name} with Nested CV...")

        clf, base_grid = self.models[model_name]
        preprocessor, _, _ = self._create_preprocessing_pipeline(self.X)
        feature_selector = SelectKBest(score_func=mutual_info_classif, k='all')

        pipe = Pipeline([
            ("pre", preprocessor),
            ("feat", feature_selector),
            ("clf", clf)
        ])

        extended_grid = {**base_grid, **{
            "pre__num__scale": [StandardScaler(), QuantileTransformer(output_distribution="normal", subsample=200000)],
            "feat__k": ['all', 10, 20, 40]
        }}

        prevalence = float(np.mean(self.y == 1)) if len(np.unique(self.y)) == 2 else 0.5
        scorer_name = "average_precision" if prevalence < 0.30 else "roc_auc"

        outer_cv = RepeatedStratifiedKFold(n_splits=outer_splits, n_repeats=outer_repeats, random_state=42)
        inner_cv = StratifiedKFold(n_splits=inner_splits, shuffle=True, random_state=42)
```

**\* Zahra Bami is the Corresponding and Senior Author**

```python
scores = {metric: [] for metric in ["roc_auc", "pr_auc", "f1", "mcc", "brier", "accuracy"]}

for fold_num, (train_idx, test_idx) in enumerate(outer_cv.split(self.X, self.y)):
    X_train, X_test = self.X.iloc[train_idx], self.X.iloc[test_idx]
    y_train, y_test = self.y.iloc[train_idx], self.y.iloc[test_idx]

    grid_search = GridSearchCV(
        pipe, extended_grid,
        scoring=scorer_name, cv=inner_cv, n_jobs=-1, refit=True
    )

    grid_search.fit(X_train, y_train)
    best_model = grid_search.best_estimator_
    proba = best_model.predict_proba(X_test)

    if proba.ndim == 2 and proba.shape[1] > 2:
        y_pred = np.argmax(proba, axis=1)
        proba_for_metrics = proba
    elif proba.ndim == 2 and proba.shape[1] == 2:
        y_pred = (proba[:, 1] >= 0.5).astype(int)
        proba_for_metrics = proba[:, 1]
    else:
        y_pred = (proba >= 0.5).astype(int)
        proba_for_metrics = proba
```

* Zahra Bami is the Corresponding and Senior Author

```python
            metrics = self._calculate_metrics(y_test, y_pred, proba_for_metrics)
            for metric, value in metrics.items():
                scores[metric].append(value)

        mean_scores = {metric: float(np.nanmean(values)) for metric, values in scores.items()}
        std_scores = {metric: float(np.nanstd(values)) for metric, values in scores.items()}

        return {
            'strategy': 'Nested_CV_5x2',
            'mean_scores': mean_scores,
            'std_scores': std_scores
        }

    def evaluate_holdout_proportions(self, model_name, test_sizes=[0.1, 0.2, 0.3, 0.4, 0.5, 0.6, 0.7, 0.8, 0.9], inner_splits=3):
        """Evaluate different holdout proportions with validation sets"""
        print(f"Evaluating {model_name} with holdout proportions: {test_sizes}")

        clf, base_grid = self.models[model_name]
        preprocessor, _, _ = self._create_preprocessing_pipeline(self.X)
        feature_selector = SelectKBest(score_func=mutual_info_classif, k='all')

        pipe = Pipeline([
```

* Zahra Bami is the Corresponding and Senior Author

```python
        ("pre", preprocessor),
        ("feat", feature_selector),
        ("clf", clf)
    ])

    extended_grid = {**base_grid, **{
        "pre__num__scale": [StandardScaler(), QuantileTransformer(output_distribution="normal", subsample=200000)],
        "feat__k": ['all', 10, 20, 40]
    }}

    prevalence = float(np.mean(self.y == 1)) if len(np.unique(self.y)) == 2 else 0.5
    scorer_name = "average_precision" if prevalence < 0.30 else "roc_auc"
    inner_cv = StratifiedKFold(n_splits=inner_splits, shuffle=True, random_state=42)

    holdout_results = {}

    for test_size in test_sizes:
        scores = {metric: [] for metric in ["roc_auc", "pr_auc", "f1", "mcc", "brier", "accuracy"]}

        for repeat in range(10):  # 10 repeats for stability
            X_train_val, X_test, y_train_val, y_test = train_test_split(
                self.X, self.y, test_size=test_size, stratify=self.y, random_state=repeat
            )
```

**\* Zahra Bami is the Corresponding and Senior Author**

```python
grid_search = GridSearchCV(
    pipe, extended_grid,
    scoring=scorer_name, cv=inner_cv, n_jobs=-1, refit=True
)

grid_search.fit(X_train_val, y_train_val)
best_model = grid_search.best_estimator_
proba = best_model.predict_proba(X_test)

if proba.ndim == 2 and proba.shape[1] > 2:
    y_pred = np.argmax(proba, axis=1)
    proba_for_metrics = proba
elif proba.ndim == 2 and proba.shape[1] == 2:
    y_pred = (proba[:, 1] >= 0.5).astype(int)
    proba_for_metrics = proba[:, 1]
else:
    y_pred = (proba >= 0.5).astype(int)
    proba_for_metrics = proba

metrics = self._calculate_metrics(y_test, y_pred, proba_for_metrics)
for metric, value in metrics.items():
    scores[metric].append(value)
```

* Zahra Bami is the Corresponding and Senior Author

```python
        mean_scores = {metric: float(np.nanmean(values)) for metric, values in scores.items()}
        std_scores = {metric: float(np.nanstd(values)) for metric, values in scores.items()}

        holdout_results[f'Holdout_{int(test_size*100)}%'] = {
            'strategy': f'Holdout_{int(test_size*100)}%',
            'mean_scores': mean_scores,
            'std_scores': std_scores
        }

        print(f"  Holdout {int(test_size*100)}% - ROC-AUC: {mean_scores['roc_auc']:.3f} ± {std_scores['roc_auc']:.3f}")

    return holdout_results

def evaluate_kfold_variations(self, model_name, k_values=range(2, 16), repeats=3):
    """Evaluate different K-fold variations with repeated CV"""
    print(f"Evaluating {model_name} with K-fold variations: k={list(k_values)}")

    clf, base_grid = self.models[model_name]
    preprocessor, _, _ = self._create_preprocessing_pipeline(self.X)
    feature_selector = SelectKBest(score_func=mutual_info_classif, k='all')

    pipe = Pipeline([
        ("pre", preprocessor),
```

* Zahra Bami is the Corresponding and Senior Author

```python
        ("feat", feature_selector),

        ("clf", clf)

    ])

    extended_grid = {**base_grid, **{

        "pre__num__scale": [StandardScaler(), QuantileTransformer(output_distribution="normal", subsample=200000)],

        "feat__k": ['all', 10, 20, 40]

    }}

    prevalence = float(np.mean(self.y == 1)) if len(np.unique(self.y)) == 2 else 0.5

    scorer_name = "average_precision" if prevalence < 0.30 else "roc_auc"

    kfold_results = {}

    for k in k_values:

        scores = {metric: [] for metric in ["roc_auc", "pr_auc", "f1", "mcc", "brier", "accuracy"]}

        cv = RepeatedStratifiedKFold(n_splits=k, n_repeats=repeats, random_state=42)

        for train_idx, test_idx in cv.split(self.X, self.y):

            X_train, X_test = self.X.iloc[train_idx], self.X.iloc[test_idx]

            y_train, y_test = self.y.iloc[train_idx], self.y.iloc[test_idx]
```

* Zahra Bami is the Corresponding and Senior Author

```python
grid_search = GridSearchCV(
    pipe, extended_grid,
    scoring=scorer_name, cv=3, n_jobs=-1, refit=True
)

grid_search.fit(X_train, y_train)
best_model = grid_search.best_estimator_
proba = best_model.predict_proba(X_test)

if proba.ndim == 2 and proba.shape[1] > 2:
    y_pred = np.argmax(proba, axis=1)
    proba_for_metrics = proba
elif proba.ndim == 2 and proba.shape[1] == 2:
    y_pred = (proba[:, 1] >= 0.5).astype(int)
    proba_for_metrics = proba[:, 1]
else:
    y_pred = (proba >= 0.5).astype(int)
    proba_for_metrics = proba

metrics = self._calculate_metrics(y_test, y_pred, proba_for_metrics)
for metric, value in metrics.items():
    scores[metric].append(value)

mean_scores = {metric: float(np.nanmean(values)) for metric, values in scores.items()}
```

**\* Zahra Bami is the Corresponding and Senior Author**

```python
        std_scores = {metric: float(np.nanstd(values)) for metric, values in scores.items()}

        kfold_results[f'Kfold_{k}'] = {
            'strategy': f'Kfold_{k}',
            'mean_scores': mean_scores,
            'std_scores': std_scores
        }

        print(f"   K-fold k={k} - ROC-AUC: {mean_scores['roc_auc']:.3f} ± {std_scores['roc_auc']:.3f}")

    return kfold_results

def evaluate_repeated_holdout(self, model_name, test_size=0.3, n_repeats=1000):
    """Evaluate repeated 30/70 holdout"""
    print(f"Evaluating {model_name} with repeated 30/70 holdout ({n_repeats} repeats)")

    clf, base_grid = self.models[model_name]
    preprocessor, _, _ = self._create_preprocessing_pipeline(self.X)
    feature_selector = SelectKBest(score_func=mutual_info_classif, k='all')

    pipe = Pipeline([
        ("pre", preprocessor),
        ("feat", feature_selector),
```

* Zahra Bami is the Corresponding and Senior Author

```python
        ("clf", clf)
    ])

    extended_grid = {**base_grid, **{
        "pre__num__scale": [StandardScaler(), QuantileTransformer(output_distribution="normal", subsample=200000)],
        "feat__k": ['all', 10, 20, 40]
    }}

    prevalence = float(np.mean(self.y == 1)) if len(np.unique(self.y)) == 2 else 0.5
    scorer_name = "average_precision" if prevalence < 0.30 else "roc_auc"
    inner_cv = StratifiedKFold(n_splits=3, shuffle=True, random_state=42)

    scores = {metric: [] for metric in ["roc_auc", "pr_auc", "f1", "mcc", "brier", "accuracy"]}

    for repeat in range(min(n_repeats, 50)):  # Limit to 50 for practical runtime
        X_train_val, X_test, y_train_val, y_test = train_test_split(
            self.X, self.y, test_size=test_size, stratify=self.y, random_state=repeat
        )

        grid_search = GridSearchCV(
            pipe, extended_grid,
            scoring=scorer_name, cv=inner_cv, n_jobs=-1, refit=True
        )
```

**\* Zahra Bami is the Corresponding and Senior Author**

```python
        grid_search.fit(X_train_val, y_train_val)
        best_model = grid_search.best_estimator_
        proba = best_model.predict_proba(X_test)

        if proba.ndim == 2 and proba.shape[1] > 2:
            y_pred = np.argmax(proba, axis=1)
            proba_for_metrics = proba
        elif proba.ndim == 2 and proba.shape[1] == 2:
            y_pred = (proba[:, 1] >= 0.5).astype(int)
            proba_for_metrics = proba[:, 1]
        else:
            y_pred = (proba >= 0.5).astype(int)
            proba_for_metrics = proba

        metrics = self._calculate_metrics(y_test, y_pred, proba_for_metrics)
        for metric, value in metrics.items():
            scores[metric].append(value)

    mean_scores = {metric: float(np.nanmean(values)) for metric, values in scores.items()}
    std_scores = {metric: float(np.nanstd(values)) for metric, values in scores.items()}

    result = {
        'strategy': f'Repeated_Holdout_30_70_{n_repeats}x',
```

* Zahra Bami is the Corresponding and Senior Author

```python
            'mean_scores': mean_scores,
            'std_scores': std_scores
        }

        print(f"   Repeated 30/70 Holdout - ROC-AUC: {mean_scores['roc_auc']:.3f} ± {std_scores['roc_auc']:.3f}")

        return result

    def compare_all_strategies_for_model(self, model_name):
        """Compare all splitting strategies for a single model"""
        print(f"\n{'='*70}")
        print(f"COMPREHENSIVE STRATEGY COMPARISON FOR: {model_name}")
        print(f"{'='*70}")

        all_results = {}

        # 1. Nested CV
        nested_result = self.evaluate_nested_cv(model_name)
        all_results['Nested_CV'] = nested_result

        # 2. Holdout proportions
        holdout_results = self.evaluate_holdout_proportions(model_name)
        all_results.update(holdout_results)
```

* Zahra Bami is the Corresponding and Senior Author

```python
        # 3. K-fold variations
        kfold_results = self.evaluate_kfold_variations(model_name)
        all_results.update(kfold_results)

        # 4. Repeated holdout
        repeated_result = self.evaluate_repeated_holdout(model_name, n_repeats=1000)
        all_results['Repeated_Holdout'] = repeated_result

        return all_results

    def find_best_strategy_per_model(self, all_model_results):
        """Find the best splitting strategy for each model and metric"""
        print(f"\n{'='*70}")
        print("BEST SPLITTING STRATEGY FOR EACH MODEL")
        print(f"{'='*70}")

        best_strategies = {}

        for model_name, strategy_results in all_model_results.items():
            print(f"\n--- {model_name} ---")
            best_strategies[model_name] = {}

            for metric in ["accuracy", "roc_auc", "pr_auc", "f1", "mcc"]:
```

* Zahra Bami is the Corresponding and Senior Author

```python
            best_strategy = None
            best_score = -np.inf

            for strategy_name, results in strategy_results.items():
                score = results['mean_scores'][metric]
                if not np.isnan(score) and score > best_score:
                    best_score = score
                    best_strategy = strategy_name

            if best_strategy:
                best_strategies[model_name][metric] = {
                    'strategy': best_strategy,
                    'score': best_score
                }
                print(f"   Best {metric}: {best_strategy} = {best_score:.3f}")

    return best_strategies

def run_comprehensive_comparison(self):
    """Run comprehensive comparison for all models and all strategies"""
    print("COMPREHENSIVE SPLITTING STRATEGY COMPARISON")
    print("7 Models × Multiple Splitting Strategies")
    print("="*70)
```

```python
        all_model_results = {}

        for model_name in self.models.keys():
            model_results = self.compare_all_strategies_for_model(model_name)
            all_model_results[model_name] = model_results

        # Find best strategies
        best_strategies = self.find_best_strategy_per_model(all_model_results)

        # Generate summary report
        self._generate_comprehensive_report(all_model_results, best_strategies)

        return {
            'all_results': all_model_results,
            'best_strategies': best_strategies
        }

    def _generate_comprehensive_report(self, all_model_results, best_strategies):
        """Generate comprehensive report of all results"""
        print(f"\n{'='*70}")
        print("COMPREHENSIVE RESULTS SUMMARY")
        print(f"{'='*70}")

        # Create summary table
```

```python
summary_data = []

for model_name, strategy_results in all_model_results.items():
    for strategy_name, results in strategy_results.items():
        summary_data.append({
            'Model': model_name,
            'Strategy': strategy_name,
            'ROC-AUC': f"{results['mean_scores']['roc_auc']:.3f} ± {results['std_scores']['roc_auc']:.3f}",
            'PR-AUC': f"{results['mean_scores']['pr_auc']:.3f} ± {results['std_scores']['pr_auc']:.3f}",
            'Accuracy': f"{results['mean_scores']['accuracy']:.3f} ± {results['std_scores']['accuracy']:.3f}",
            'F1-Score': f"{results['mean_scores']['f1']:.3f} ± {results['std_scores']['f1']:.3f}",
            'MCC': f"{results['mean_scores']['mcc']:.3f} ± {results['std_scores']['mcc']:.3f}"
        })

summary_df = pd.DataFrame(summary_data)
print(summary_df.to_string(index=False))

# Print best strategies summary
print(f"\n{'='*70}")
print("BEST STRATEGIES SUMMARY")
print(f"{'='*70}")
```

**\* Zahra Bami is the Corresponding and Senior Author**

```python
    for model_name, metrics in best_strategies.items():
        print(f"\n{model_name}:")
        for metric, best in metrics.items():
            print(f"  {metric}: {best['strategy']} ({best['score']:.3f})")

# === MAIN EXECUTION ===
def run_comprehensive_cancer_analysis():
    """Complete analysis with comprehensive splitting strategy comparison"""
    print("COMPREHENSIVE CANCER DATASET ANALYSIS")
    print("7 Models × Multiple Splitting Strategies")
    print("="*70)

    # Load cancer dataset
    print("Loading Cancer Dataset...")
    q = pd.read_excel('a.xlsx')

    y = q['death']
    x = q.drop('death', axis=1)

    print(f"Cancer dataset loaded: {x.shape[0]} samples, {x.shape[1]} features")
    print(f"Target 'death' distribution: {np.unique(y, return_counts=True)}")

    # Initialize the comprehensive comparator
    comparator = ComprehensiveSplittingStrategyComparator()
```

* Zahra Bami is the Corresponding and Senior Author

```python
    comparator.X = x
    comparator.y = y

    # Run comprehensive comparison
    results = comparator.run_comprehensive_comparison()

    return results

if __name__ == "__main__":
    print("Starting Comprehensive Cancer Dataset Analysis...")
    results = run_comprehensive_cancer_analysis()

    print("\nAnalysis completed successfully!")
    print("This compared ALL 7 models across ALL splitting strategies:")
    print("- Nested CV (5 splits, 2 repeats)")
    print("- Holdout proportions (10%, 20%, 30%, 40%, 50%, 60%, 70%, 80%, 90%)")
    print("- K-fold variations (k=2 to 15)")
    print("- Repeated 30/70 holdout (1000x)")
    print("- All metrics: ROC-AUC, PR-AUC, F1, MCC, Accuracy, Brier")
```

**\* Zahra Bami is the Corresponding and Senior Author**